\documentclass{article}

\usepackage[preprint]{neurips_2023}

\usepackage[utf8]{inputenc} 
\usepackage[T1]{fontenc}    
\usepackage{hyperref}       
\usepackage{url}            
\usepackage{booktabs}       
\usepackage{amsfonts}       
\usepackage{nicefrac}       
\usepackage{microtype}      
\usepackage{subcaption}

\usepackage{natbib}
\setcitestyle{numbers,square}
\usepackage{amsmath}
\usepackage[capitalize]{cleveref}
\crefname{section}{Sec.}{Secs.}
\Crefname{section}{Section}{Sections}
\Crefname{table}{Table}{Tables}
\crefname{table}{Tab.}{Tabs.}
\usepackage{graphicx} 
\usepackage{algorithm}
\usepackage{algorithmic}
\usepackage{color}
\usepackage{alltt}
\usepackage[dvipsnames]{xcolor,colortbl}
\usepackage{enumitem}
\usepackage{wrapfig}

\title{Experts Weights Averaging: A New General Training Scheme for Vision Transformers}

\author{Yongqi Huang\textsuperscript{1}$^\dagger$~,~Peng Ye\textsuperscript{1}$^\dagger$~, ~Xiaoshui Huang\textsuperscript{2}~,~Sheng Li\textsuperscript{3}~, \\ {\bfseries Tao Chen\textsuperscript{1}\thanks{Corresponding Author (eetchen@fudan.edu.cn).~~~$^\dagger$Equal Contribution.}~, ~Tong He\textsuperscript{2}, ~Wanli Ouyang\textsuperscript{2}}\\\\
\textsuperscript{1}School of Information Science and Technology, Fudan University, 
\textsuperscript{2}Shanghai AI Lab,\\ 
\textsuperscript{3}Jiangxi University of Finance and Economics\\
{\tt\small \{19307130163, yepeng20\}@fudan.edu.cn}}

\begin{document}

\maketitle

\begin{abstract}
Structural re-parameterization is a general training scheme for Convolutional Neural Networks (CNNs), which achieves performance improvement without increasing inference cost. 
As Vision Transformers (ViTs) are gradually surpassing CNNs in various visual tasks, one may question: \textit{if a training scheme specifically for ViTs
exists that can also achieve performance improvement without increasing inference cost?} Recently, Mixture-of-Experts (MoE) has attracted increasing attention, as it can efficiently scale up the capacity of Transformers at a fixed cost through sparsely activated experts. Considering that MoE can also be viewed as a multi-branch structure, \textit{can we utilize MoE to implement a ViT training scheme similar to structural re-parameterization?} In this paper, we affirmatively answer these questions, with a new general training strategy for ViTs. Specifically, we decouple the training and inference phases of ViTs.During training, we replace some Feed-Forward Networks (FFNs) of the ViT with specially designed, more efficient MoEs that assign tokens to experts by random uniform partition, and perform Experts Weights Averaging (EWA) on these MoEs at the end of each iteration. After training, we convert each MoE into an FFN by averaging the experts, transforming the model back into original ViT for inference. We further provide a theoretical analysis to show why and how it works. Comprehensive experiments across various 2D and 3D visual tasks, ViT architectures, and datasets validate the effectiveness and generalizability of the proposed training scheme. Besides, our training scheme can also be applied to improve performance when fine-tuning ViTs. Lastly, but equally important, the proposed EWA technique can significantly improve the effectiveness of naive MoE in various 2D visual small datasets and 3D visual tasks.
\end{abstract}

\section{Introduction}
\label{sec:intro}
In recent years, the development of general training schemes has played a critical role in deep learning. These methods enhance existing models without modifying their architectures, and do not require any assumptions about specific architectures. As a result, they can be employed for various models and provide uniform benefits.

\begin{figure}[t]
  \centering
  \includegraphics[width=\linewidth]{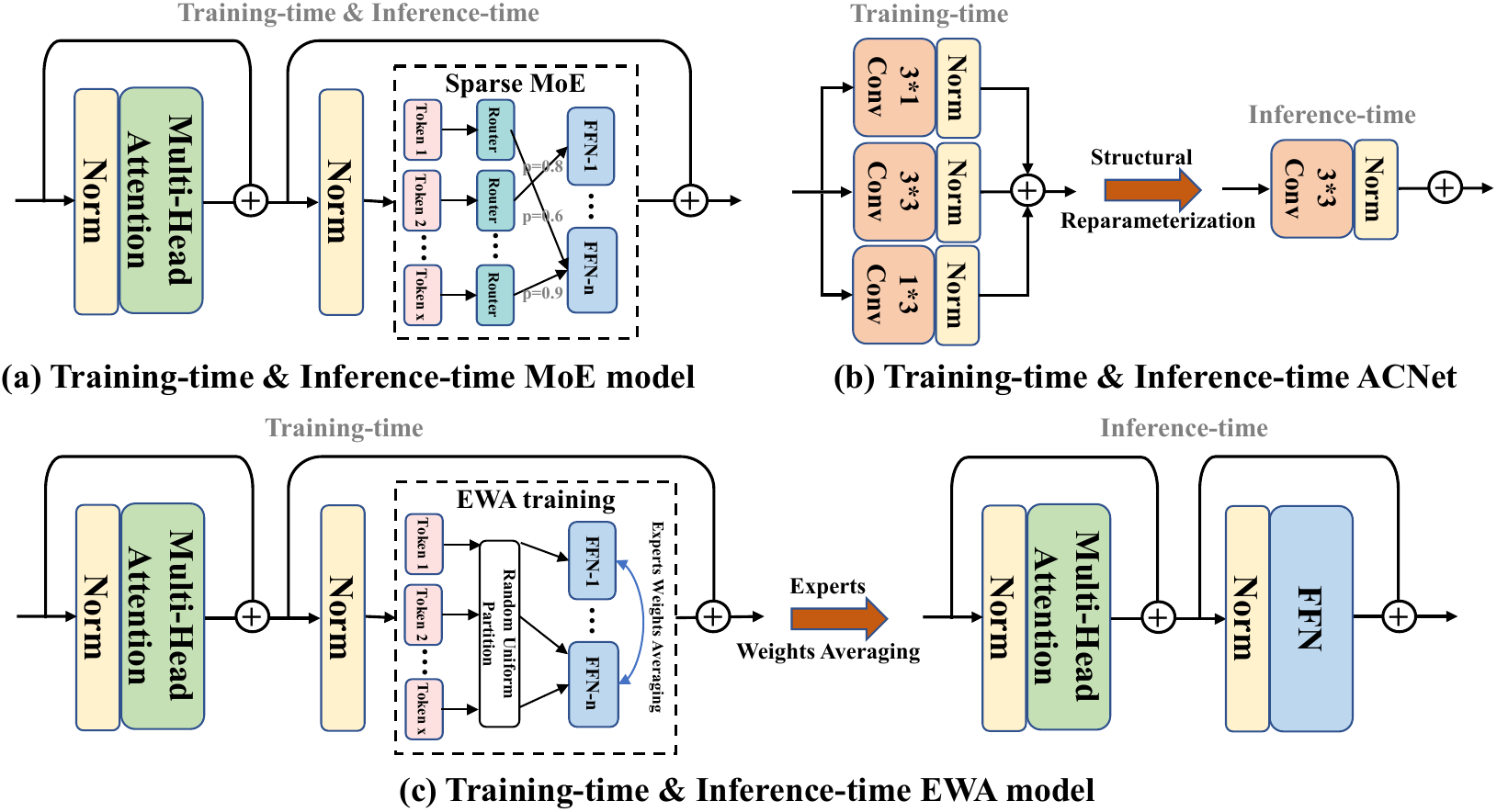}
  \vspace{-5mm}
  \caption{Training-time \& inference-time comparison of (a) recent MoE block, (b) typical structural re-parameterization block, and (c) our EWA training block. Each FFN denotes an expert of MoE.}
  \label{fig:EWA}
  \vspace{-5mm}
\end{figure}

For CNN architecture, the widely-used general training scheme is structural re-parameterization~\cite{ding2019acnet,ding2021diverse,ding2021repvgg,ding2021repmlp} (Fig.~\ref{fig:EWA}(b)). This approach decouples the training and inference stages of the CNN model by using a multi-branch structure during training, which is then equivalently converted back to the original single-branch structure during inference. Structural re-parameterization achieves improved performance without increasing inference cost. However, the multi-branch structure used during training may slow down the training speed (as shown in the second column of Tab.~\ref{tab:ImgReg}, its training latency is up to 2.18$\times$ of vanilla training).
Additionally, structural re-parameterization only applies to CNNs, not to other recent stronger architectures like ViTs (see the sixth column of Tab.~\ref{tab:ImgReg}). 

\begin{table}[t]
	\centering
        \caption{
        The comparison of achieved effects between vanilla training, structural reparam., MoE training, and our EWA training. The testbed is a GTX3090.
        For structural reparam., the baseline is ResNet-18, and the comparison model is its ACNet counterpart~\cite{ding2019acnet}. 
        For MoE training, the baseline is standard ViT-S, and the comparison model is our V-MoE-S which replaces every other FFN of ViT-S with an 8-expert MoE using top-1 routing~\cite{fedus2022switch}. For EWA training, we replace top-1 routing with random uniform partition for V-MoE-S.
        To test latency, we use 128$\times$3$\times$224$\times$224 as the input.}
        \vspace{1mm}
	\begin{tabular}{ccccccc}
\hline
-                                                               & \begin{tabular}[c]{@{}c@{}}Training\\ Latency\end{tabular} & \begin{tabular}[c]{@{}c@{}}Training\\ Params\end{tabular} & \begin{tabular}[c]{@{}c@{}}Inference\\ Latency\end{tabular} & \begin{tabular}[c]{@{}c@{}}Inference\\ Params\end{tabular} & Architecture & Applications      \\ \hline
\begin{tabular}[c]{@{}c@{}}Vanilla\\ Training\end{tabular}      & 1.0$\times$                                                          & 1.0$\times$                                                         & 1.0$\times$                                                              & 1.0$\times$                                                             & -            & -                 \\ \hline
\begin{tabular}[c]{@{}c@{}}Structural\\ Reparam.\end{tabular}   & 2.18$\times$                                                         & 1.63$\times$                                                        & 1.0$\times$                                                              & 1.0$\times$                                                             & CNN          & 2D                \\ \hline
\begin{tabular}[c]{@{}c@{}}MoE\\ Training\end{tabular}          & 1.12$\times$                                                         & 3.25$\times$                                                        & 1.12$\times$                                                             & 3.25$\times$                                                            & ViT          & 2D                \\ \hline
\textbf{\begin{tabular}[c]{@{}c@{}}EWA\\ Training\end{tabular}} & \textbf{1.07$\times$}                                                & \textbf{3.25$\times$}                                               & \textbf{1.0$\times$}                                                     & \textbf{1.0$\times$}                                                    & \textbf{ViT} & \textbf{2D \& 3D} \\ \hline
\end{tabular}
	\label{tab:ImgReg}
\vspace{-4mm}
\end{table}

For transformer architecture, the recent general training method involves replacing the feed-forward network (FFN) layer with a sparse mixture of experts (MoE)~\cite{lepikhin2020gshard,fedus2022switch,du2022glam,riquelme2021scaling,hwang2022tutel} (Fig.~\ref{fig:EWA}(a)). The MoE is a novel network component that includes multiple experts (FFNs) with unique weights and a trainable routing network. During the training and inference stages, the routing network selects a sparse combination of experts for each input, enabling efficient scaling of the capacity of transformer models through sparse computation. In the field of natural language processing, many sparse MoE
models~\cite{lepikhin2020gshard,fedus2022switch,du2022glam} have demonstrated that they outperform their dense counterparts. 

However, MoE still has some limitations that restrict its widespread use in Vision Transformers (ViTs).
Firstly, V-MoE~\cite{riquelme2021scaling} and SwinV2-MoE~\cite{hwang2022tutel} were pre-trained on large-scale datasets (JFT-300M~\cite{sun2017revisiting}, ImageNet-22K~\cite{ridnik2021imagenet}), and when pre-trained on ImageNet-1K~\cite{deng2009imagenet}, V-MoE's performance is lower than ViT~\cite{xue2022one}. Furthermore, there is no research studying MoE Vision Transformer on smaller datasets such as CIFAR-100~\cite{krizhevsky2009learning} and Tiny-ImageNet~\cite{le2015tiny}. Secondly, current MoE ViTs~\cite{riquelme2021scaling,xue2022go,xue2022one,hwang2022tutel,komatsuzaki2022sparse} are focused on 2D visual tasks, and there are no MoE ViTs for 3D visual tasks (see the fourth row of Tab.~\ref{tab:ImgReg}). Thirdly, the top-k routing mechanism of MoE itself is not efficient and effective. As shown in the fourth row of Tab.~\ref{tab:ImgReg}, the training and inference latency both increase to 1.12$\times$ since top-k routing requires additional computation to make routing decisions for each input, and the training and inference parameters both largely increase to 3.25$\times$ that burdens model employing.
Besides, top-k routing easily leads to load imbalance across experts, thus many studies are devoted to handling this problem through various designs, such as adding auxiliary load balancing losses~\cite{lepikhin2020gshard,fedus2022switch},
making routing decisions globally~\cite{lewis2021base}, or redesigning the routing algorithm~\cite{,roller2021hash,zhou2022mixture}.

Therefore, the following questions arise: \textit{Is there a new training method that can efficiently train the sparsely activated experts during the training phase, and can a sparse model be converted to a dense model without additional computation cost during the inference phase? Can we further extend the effectiveness of MoE to both 2D visual datasets and 3D visual tasks?}

In this paper, we design a new general training strategy for ViTs called Experts Weights Averaging (EWA) training. Our approach achieves performance improvement without increasing inference latency and parameters,
as shown in the second and last rows of Tab.~\ref{tab:ImgReg}. In detail, we decouple the training and inference stages of ViT, as shown in Fig.~\ref{fig:EWA}(c). 
During training, we replace some FFNs of vanilla ViT with specially designed, more efficient MoEs that assign tokens to experts by Random Uniform Partition (RUP), requiring no additional parameters, special designs, or auxiliary losses. Further, at the end of each iteration, we perform Experts Weights Averaging (EWA) on each MoE.
During inference, by simply averaging the experts of each MoE into a single FFN, we can convert the sparse MoE model into a vanilla ViT model without any performance loss. 


We conduct comprehensive experiments on various 2D and 3D visual tasks, ViT architectures, and datasets to validate the effectiveness and generalization of the proposed training scheme. {For example, our proposed training method achieves 1.72\% improvement on the image classification task and 1.74\%  mIoU improvement on point cloud semantic segmentation task.}
Furthermore, EWA training can also be used for finetuning pretrained ViT models. {Our training method improves the ViT-B finetuning on CIFAR100 from 90.71\% to 91.42\% ($\uparrow0.71\%$).}
Finally, we find that the key technology of Experts Weights Averaging in EWA training can greatly improve the effectiveness of naive MoE in various 2D visual small datasets and 3D visual tasks. 

The main contributions are summarized as:
\begin{itemize}[leftmargin=*]
\vspace{-3pt}
\item We propose a new general training scheme for ViTs, achieving performance improvement without increasing inference latency and parameters.
During training, we replace some of the FFNs in the ViT architecture with MoE using Random Uniform Partition and perform Experts Weights Averaging on these MoE layers after each model weight update. After training, we convert each MoE layer in the model into an FFN layer by averaging the experts, thus transforming the model back to the original ViT architecture for inference.
\item We validate that EWA training can bring unified performance improvement to various Vision Transformer architectures on different 2D/3D visual tasks and datasets. Furthermore, we find that EWA training can also be used for fine-tuning pretrained ViT models, achieving even further performance improvement. We provide theoretical analysis to show why and how it works.
\item We explore the effectiveness of naive MoE on small 2D visual datasets and 3D visual tasks. We find that using naive MoE results in lower performance than dense counterparts, but the proposed Experts Weights Averaging technique can greatly improve the effectiveness of naive MoE.
\vspace{-3pt}
\end{itemize}

\section{Related works}
\subsection{MoE} \vspace{-1mm}
The Sparsely-Gated Mixture-of-Experts (MoE) is first introduced in~\cite{shazeer2017outrageously} and has shown significant improvements in training time, model capacity, and performance. 
Furthermore, when MoE is introduced into transformer-based language models, it receives increasing attention~\cite{lepikhin2020gshard,fedus2022switch,du2022glam}. To date, many works have improved MoE's key component, that is the routing network~\cite{fedus2022switch,lewis2021base,roller2021hash,zhou2022mixture,zuo2021taming}. On the one hand, for traditional learnable routing networks, Switch Transformer~\cite{fedus2022switch} simplifies routing by selecting only the top-1 expert for each token, Base Layer~\cite{lewis2021base} handles a linear assignment problem to perform routing, and Expert Choice~\cite{zhou2022mixture} routes top-k tokens to each expert. On the other hand, Hash Layer~\cite{roller2021hash} designs deterministic hash for token routing, and THOR~\cite{zuo2021taming} randomly activates experts for each input.
In addition to routing network improvements, many works have explored MoE's implementation on modern hardwares~\cite{he2021fastmoe,rajbhandari2022deepspeed,nie2022hetumoe,hwang2022tutel,nie2023flexmoe}, scaling properties~\cite{fedus2022switch,artetxe2021efficient,du2022glam,clark2022unified}, and applications~\cite{lepikhin2020gshard,riquelme2021scaling,you2021speechmoe,mustafa2022multimodal,xue2022go,xue2022one,komatsuzaki2022sparse,zhang2022mixture,chen2022mod}. However, most of MoE works are based on transformer-based language models, and only a few MoE Vision Transformer works exist in the computer vision field~\cite{riquelme2021scaling,xue2022go,xue2022one,hwang2022tutel,komatsuzaki2022sparse}. Riquelme et al.~\cite{riquelme2021scaling} create V-MoE by replacing some FFN layers of ViT with MoE layers, pre-train V-MoE on JFT-300M or JFT-3B and achieve great success in image classification. Hwang et al.~\cite{hwang2022tutel} 
add MoE layers to the Swin-transformer V2 architecture for pre-training on ImageNet-22K for image classification and object detection tasks. Xue et al.~\cite{xue2022go} propose a more parameter-efficient and effective framework, WideNet, by replacing FFN with MoE and sharing the MoE layer across transformer blocks, achieving performance improvement on ImageNet-1K pre-training. Komatsuzaki et al.~\cite{komatsuzaki2022sparse} use a pre-trained dense transformer checkpoint to warm-start the training of a MoE model, and achieve promising results on JFT-300M pre-training and ImageNet-1K fine-tuning. Xue et al.~\cite{xue2022one} achieve knowledge distillation from a pre-trained MoE model to learn a dense model, and verify it on ImageNet-1K. 

In this paper, we propose a new general training for improving the performance of ViTs without increasing inference cost, which has different targets and implementation methods from MoE. Besides, previous MoE-ViT studies have primarily focused on larger 2D visual datasets, with little exploration conducted on smaller 2D visual datasets and 3D vision. This paper shows that adding naive MoE to ViT degrades the performance of ViTs on smaller 2D visual datasets and 3D vision, and introduces a novel Early EMA training approach that can greatly improve their performance.
\vspace{-1mm}
\subsection{Structural Re-parameterization} \vspace{-1mm}
The essence of structural re-parameterization is to use a multi-branch structure instead of a convolution or fully-connected layer to enhance the model during training. After training, multiple branches are fused for inference using equivalent parameter transformations or specific algorithms. The typical examples are ACNet~\cite{ding2019acnet}, DBB~\cite{ding2021diverse} and RepVGG~\cite{ding2021repvgg}. Specifically, ACNet~\cite{ding2019acnet} replaces common $K\times K$ convolutions in CNN with ACBlocks that have parallel structures of $K\times1$, $1\times K$, and $K\times K$ convolutions during training. After training, ACBlocks are equivalently converted to a single $K\times K$ convolution layer. DBB~\cite{ding2021diverse} uses a multi-branch structure of $1\times1$ convolution, $K\times K$ convolution, $1\times1$-$K\times K$ convolution, and $1\times1$ convolution-average pooling to replace the original $K\times K$ convolution during training, which is equivalently converted to a single $K\times K$ convolution during inference. 
RepVGG~\cite{ding2021repvgg} adds parallel $1\times1$ convolution branch and identity branch to each $3\times3$ convolution layer during training and is equivalently converted to a single $3\times3$ convolution layer after training. Due to the performance improvement without increasing inference cost it brings, structural re-parameterization has received increasing attention and application in various computer vision tasks. For example, ExpandNets~\cite{guo2020expandnets} uses it to design compact models, RepNAS~\cite{zhang2021repnas} uses it in neural architecture search, and ResRep~\cite{ding2021resrep} combines it with pruning. 

However, previous studies on structural re-parameterization have mainly focused on CNN architectures, making them incompatible with ViT architectures. 
This paper introduces a novel approach that decouples the training and inference stages of ViT architectures. During training, we use efficient MoE sparse computation with Random Uniform Partition and Experts Weights Averaging mechanisms. During inference, we convert the model to the original ViT architecture, resulting in performance improvements without increasing inference parameters and latency. This approach is the first of its kind to apply MoE sparse computation to improve the performance of ViT architectures.

\vspace{-1mm}
\subsection{Weight Averaging} \vspace{-1mm}
Weight averaging is a commonly used technique in deep learning and has been widely used in various tasks. For instance, in self-supervised and semi-supervised learning, \cite{tarvainen2017mean,caron2021emerging} use the exponential moving average (EMA) weights of the student model as the teacher to provide a smoother target for the student. Similarly, in online knowledge distillation,~\cite{wu2021peer} employs the EMA weights of each branch as the peer mean teacher to collaboratively transfer knowledge among branches. Additionally, many studies use weight averaging to enhance the model's generalization ability~\cite{izmailov2018averaging,wortsman2022robust,wortsman2022model,rame2022diverse}. For example, stochastic weight averaging (SWA)~\cite{izmailov2018averaging} averages the weights of multiple points on the SGD optimization trajectory and achieves better generalization ability than conventional training. Wortsman et al. \cite{wortsman2022robust} average the original zero-shot model and the fine-tuned model, resulting in improvements in both in-distribution and out-of-distribution generalization. Model Soup~\cite{wortsman2022model} averages models with the same pre-trained initialization but different hyperparameter configurations during fine-tuning to obtain a better model. DiWA~\cite{rame2022diverse} provides theoretical analysis for the success of weight averaging in OOD and proposes to reduce the covariance between predictions and improve OOD generalization by averaging the weights of diverse independently trained models. 

Unlike the studies mentioned above that conduct weight averaging at the model level to obtain a single model, our approach performs weight averaging at the expert level in MoE during the training process. By averaging the weights of other experts onto each expert, we obtain multiple experts.

\subsection{Dropout and Other variants for ViT} \vspace{-1mm}
Dropout~\cite{srivastava2014dropout} is a common technique used to improve the network generalization ability by randomly dropping neurons and their corresponding connections during training to prevent overfitting. 
For ViTs, most works use dropout with fixed drop rate for the self-attention or FFN layers. Meanwhile, Stochastic Depth~\cite{huang2016deep} is often used in ViTs to randomly drop some blocks and applies higher drop rates to deeper blocks to improve generalization. Recently, a new dropout variant for ViT called dropkey is introduced. Dropkey~\cite{li2022dropkey} assigns an adaptive operator to each attention block by randomly dropping some keys during attention computation. This strategy constrains the attention distribution, making it smoother and more effective in alleviating overfitting. Moreover, dropkey gradually reduces the drop rate as the number of blocks increases, avoiding the instability that can result from a constant drop rate in traditional dropout methods during the training process.

In our method, for each MoE layer, we randomly and evenly divide all tokens among all experts during training. This is equivalent to randomly dropping a large number of tokens at each expert. Such an operation may help prevent overfitting while ensuring the training latency remains almost equivalent to vanilla training.

\section{Method}
\subsection{PRELIMINARY} \vspace{-1mm}
{\bfseries MoE.} As shown in Fig.~\ref{fig:EWA}(a), a standard MoE layer consists of a set of $N$ experts (i.e., $N$ FFNs) $\left\{E_{1}, E_{2}, \ldots, E_{N}\right\}$, and a routing network ${G}$ with weight ${W_g}$. Given an input example ${x}$, the output of both the training-time and inference-time MoE layer can be written as:
\begin{align}
\label{formulation:moe1}&y= \sum\limits_{i=1}^NG(x)_{i} \cdot E_{i}(x) \\
\label{formulation:moe1}&G(x)=TopK(softmax(x \cdot W_{g}), k)
\end{align}
where $G(x)_{i}$ denotes the routing score to the $i$-th expert, $E_{i}(x)$ stands for the output of $i$-th expert, $TopK(,k)$ means to select top $k$ experts and set $G(x)$ of other experts as 0. Usually, $k \ll N$, which means $G(x)$ a sparse $N$-dimensional vector. When $G(x)_{i}=0$, $E_{i}(x)$ does not need to be computed.\\

{\bfseries Structural re-parameterization.} As shown in Fig.~\ref{fig:EWA}(b), structural re-parameterization employs a multi-branch CNN structure to replace the common convolutional layer during training. 
Let us denote $\left\{f_{1}, f_{2}, \ldots, f_{M}\right\}$ as the total $M$ branches (usually $M$ different operators with compatible sizes). For a given input $x$, the output of training-time structure can be expressed as:
\begin{align}
\label{formulation:reparam}
    y= \sum\limits_{i=1}^M f_{i}(x)
\end{align}
After training, these $M$ branches will be equivalently converted to a single convolutional layer $F$ for inference. For an inference-time input $x$, the output becomes $y=F(x)$.

\subsection{A New General Training Scheme} \vspace{-1mm}
\subsubsection{Motivation} \vspace{-1mm}
This paper aims to design a new general training scheme for ViTs, achieving performance improvement without increasing inference cost. As shown in Eq.~\ref{formulation:reparam}, classical structural re-parameterization only applies to CNNs composed of convolutions and significantly increases the training cost due to the multi-branch structure. As shown in Eq.~\ref{formulation:moe1}, recent MoE consists of $N$ experts and a routing network. First, the routing network brings additional parameters needed to learn and easily leads to the load imbalance problem. Second, $N$ experts largely increase the burden of deploying the model. Third, the routing score $G(x)$ is a bottleneck in converting MoE into FFN. Considering these, we first propose a specially designed, more efficient MoE to replace some FFNs of ViT for training, which assigns tokens to experts by Random Uniform Partition. Then, we propose a simple-but-effective technique called Experts Weights Averaging, which averages the weights of other experts to each expert at the end of each training iteration. As such, after training, we can convert each MoE to a single FFN for inference. We show how to apply it during training and fine-tuning below.

\subsubsection{EWA Training} \vspace{-1mm}
{\bfseries Overall.} When applying EWA training to train a ViT model from scratch, we first create a randomly initialized MoE ViT by replacing some FFNs with Random Uniform Partition based MoE layers. During training, we perform Experts Weights Averaging with share rate $\beta$ for each MoE layer after each weight update, and the $\beta$ increases linearly with the training epoch from 0 to the hyperparameter $share\ rate$. After training, we convert each MoE into an FFN by averaging the experts.

{\bfseries MoE layer with Random Uniform Partition.} Given a MoE layer that consists of $N$ experts (namely $N$ identical FFNs), $\left\{E_{1}, E_{2}, \ldots, E_{N}\right\}$, assuming these are $T$ inputs $\left\{x_{1}, x_{2}, \ldots, x_{T}\right\}$ and $T$ can be divisible by $N$. The outputs of MoE layer with Random Uniform Partition (RUP) can be written as:
\begin{align}
\label{formulation:rup1}&\left\{x_{1}, x_{2}, \ldots, x_{T}\right\} \to \left\{X_{1}, X_{2}, \ldots, X_{N}\right\} \\
\label{formulation:rup2}&y=E_{i}(x)\hspace{2em}if\,\,x \in X_{i}
\end{align}
As shown in Eq.~\ref{formulation:rup1}, the $T$ inputs are randomly and uniformly partitioned into $N$ parts, where $X_{i}$ represents the $\frac{T}{N}$ inputs that assigned to the $i$-th expert. Further, as shown in Eq.~\ref{formulation:rup2}, for each input in ${X_{i}}$, its output is $E_{i}(x)$. \textbf{We show the pytorch-like code of RUP in Appendix}.\\

{\bfseries Experts Weights Averaging.} Given the weights of $N$ experts $\left\{W_{1}, W_{2}, \ldots, W_{N}\right\}$ and share rate $\beta$, Experts Weights Averaging (EWA) performed on each MoE layer can be formulated as:
\begin{align}
\label{formulation:ewa}\overline{W_{i}}=(1-\beta)W_{i} +\sum\limits_{j \neq i}^N \frac{\beta}{N-1} W_{j}
\end{align}
$\overline{W_{i}}$ denotes the new weight of the $i$-th expert. In short, we average the weights of other experts to each expert and obtain multiple experts. \textbf{The pytorch-like code of EWA is provided in Appendix}.

{\bfseries Convert MoE into FFN.} After training, each MoE layer will be converted into an FFN layer by simply averaging the experts weights. Given a MoE layer consisting of $N$ experts $\left\{E_{1}, E_{2}, \ldots, E_{N}\right\}$, the corresponding inference-time FFN can be expressed as:
\begin{align}
FFN=\frac{1}{N}\sum\limits_{i=1}^NE_{i}
\end{align}
By this way, the training-time MoE model will be converted into a vanilla ViT model for inference. \textbf{We conduct experiments to show the rationality in Appendix}.

\subsubsection{EWA Fine-tuning} \vspace{-1mm}
{\bfseries Overall.} When applying EWA training to fine-tune a pre-trained ViT model, the only difference from training ViT from scratch is the weight initialization of the created MoE ViT model. Rather than random initialization, we initialize the MoE ViT with the given ViT model checkpoint. When replacing some FFNs with MoE layers, each expert of a specific MoE layer is initialized as a copy of the corresponding FFN layer in the pre-trained ViT model. For other layers that remain unchanged, their weights are directly inherited from the original ViT checkpoint. 

\section{Theoretical Analysis} 
In this section, we theoretically analyze \textit{why and how EWA training works}. For the convenience of analysis, we focus on one MoE layer and $m$ training steps. Assuming that the current training step is $t$,
following Eq.~\ref{formulation:ewa}, the new weight of $i$-th expert after experts weights averaging can be written as:
\begin{align}
\label{formulate:t} \overline{W_{i}^{t}}=&(1-\beta) W_{i}^{t}+\sum_{j \neq i}^N  \frac{\beta}{N-1} W_{j}^{t}
\end{align}
Then, the $(t+1)$-th training step begins, and the weights of each expert of MoE are updated as $\{W_{1}^{t+1}, W_{2}^{t+1}, \ldots, W_{N}^{t+1}\}$, where $W_{i}^{t+1}=\overline{W_{i}^{t}}+\eta \nabla \overline{W_{i}^{t}}$, $\eta$ denotes the learning rate.
Further, the experts weights averaging is performed, as shown in Eq.~\ref{formulate:t+1}. Based on Eq.~\ref{formulate:t} and $W_{i}^{t+1}$, Eq.~\ref{formulate:t+1} can be further reformulated as Eq.~\ref{formulate:t+1-1}.
\begin{align}
\label{formulate:t+1} \overline{W_{i}^{t+1}}&=(1-\beta) W_{i}^{t+1}+\sum_{j \neq i}^N  \frac{\beta}{N-1} W_{j}^{t+1} \\
\label{formulate:t+1-1}&=(1-\beta)^{2} W_{i}^{t}+\eta(1-\beta) \nabla \overline{W_{i}^{t}}+\sum_{j \neq i}^N \left[\frac{\beta}{N-1} W_{j}^{t+1}+\frac{\beta(1-\beta)}{N-1} W_{j}^{t}\right]
\end{align}
Similarly, we can get $\overline{W_{i}^{t+2}}$, $\overline{W_{i}^{t+3}}$ and $\overline{W_{i}^{t+m}}$. Through mathematical induction, we get Eq.~\ref{formulate:t+m-1}.
\begin{align}
&\label{formulate:t+m} \overline{W_{i}^{t+m}}=(1-\beta) W_{i}^{t+m}+\sum_{j \neq i}^N  \frac{\beta}{N-1} W_{j}^{t+m} \\
&\label{formulate:t+m-1}=(1-\beta)^{m+1} W_{i}^{t}+\eta \sum_{k=1}^{m}(1-\beta)^{k} \nabla \overline{W_{i}^{t+m-k}}+\frac{\beta}{N-1} \sum_{j \neq i}^N  \sum_{k=0}^{m}(1-\beta)^{m-k} \cdot W_{j}^{t+k}
\end{align}
According to the above derivation, there are two findings: 1) Observing the first term of Eq.~\ref{formulate:t+1} and Eq.~\ref{formulate:t+m}, \textit{there is a layer-wise weight decay when performing experts weights averaging}; 2) Observing the last term of Eq.~\ref{formulate:t+1-1} and Eq.~\ref{formulate:t+m-1}, \textit{there continuously aggregates multi-experts historical exponential average weights along the training iteration of EWA training}. \textbf{More Details are shown in Appendix.}

\section{Experiments}
When applying EWA training to train or fine-tune ViTs, we always use random uniform partition based MoE to replace some FFNs. Unless otherwise specified, the number of each MoE's experts is set to 4 by default. The hyperparameter $share\ rate$ is simply adjusted within \{0.1, 0.2, 0.3, 0.4, 0.5\} for different ViT architectures to get the best performance. Following V-MoE \cite{riquelme2021scaling}, where to insert these MoE layers is simply adjusted within \{every-2, last-4\}.
\textbf{All details are shown in Appendix.} 


\begin{table*}[t]
\caption{Performance comparison between vanilla and EWA training on various architectures and datasets. For ViT architectures (-XT, -XS, -l2, -T) and training methods (+SL) specifically designed for small datasets~\cite{lee2021vision}, the addition of our EWA method can still improve their performance.}
\vspace{-1mm}
\label{table:2d training}
    \centering
	\begin{minipage}{0.48\linewidth}
		\centering
        \subcaption{on CIFAR-100 dataset}
\begin{tabular}{cccc}
\hline
Model       & \begin{tabular}[c]{@{}c@{}}Vanilla\\ Top-1\end{tabular} & \begin{tabular}[c]{@{}c@{}}EWA\\ Top-1\end{tabular} & \begin{tabular}[c]{@{}c@{}}$\Delta$\\ Top-1\end{tabular} \\ \hline
ViT-S       & 72.61                                                   & 74.33                                               & +1.72                                            \\
ViT-XT      & 73.93                                                   & 75.20                                               & +1.27                                               \\
ViT-XT + SL & 77.04                                                   & 77.94                                               & +0.90                                               \\ \hline
PiT-XS      & 74.99                                                   & 75.65                                               & +0.66                                               \\
PiT-XS + SL & 79.73                                                   & 80.39                                               & +0.66                                               \\ \hline
T2T-12      & 77.19                                                   & 77.80                                               & +0.61                                               \\
T2T-12 + SL & 78.35                                                   & 78.81                                               & +0.46                                               \\ \hline
Swin-T      & 77.38                                                   & 77.65                                               & +0.27                                              \\
Swin-T + SL & 80.31                                                   & 80.52                                               & +0.21                                              \\ \hline
\end{tabular}
	\end{minipage}
	\hfill
	\begin{minipage}{0.48\linewidth}
		\centering
        \subcaption{on Tiny-ImageNet dataset}
\begin{tabular}{cccc}
\hline
Model       & \begin{tabular}[c]{@{}c@{}}Vanilla\\ Top-1\end{tabular} & \begin{tabular}[c]{@{}c@{}}EWA\\ Top-1\end{tabular} & \begin{tabular}[c]{@{}c@{}}$\Delta$\\ Top-1\end{tabular} \\ \hline
ViT-S       & 57.41                                                   & 59.50                                               & +2.09                                               \\
ViT-XT      & 55.92                                                   & 57.86                                               & +1.94                                               \\
ViT-XT + SL & 59.72                                                   & 60.75                                               & +1.03                                              \\ \hline
PiT-XS      & 59.03                                                   & 61.61                                               & +2.58                                               \\
PiT-XS + SL & 63.11                                                   & 64.53                                               & +1.42                                               \\ \hline
T2T-12      & 60.39                                                   & 61.65                                               & +1.26                                               \\
T2T-12 + SL & 62.57                                                   & 63.36                                               & +0.79                                               \\ \hline
Swin-T      & 59.70                                                   & 60.66                                               & +0.96                                               \\
Swin-T + SL & 64.34                                                   & 65.13                                               & +0.79                                               \\ \hline
\end{tabular}
\end{minipage}
\vspace{-3mm}
\end{table*}

\subsection{EWA Training} \vspace{-1mm}
\subsubsection{2D Classification on various architectures and datasets} \vspace{-1mm}
{\bfseries Settings.} We conduct comprehensive experiments on various architectures and datasets to evaluate the EWA training scheme. For datasets, we employ CIFAR-100 and Tiny-ImageNet. For architectures, besides standard ViT-S, we also adopt various ViT architectures specifically designed for small datasets such as smaller ViT-Tiny (we call it ViT-XT), PiT-XS, T2T-12, Swin-T, and their SL enhanced counterparts~\cite{lee2021vision}. SL means to apply shifted patch tokenization and locality self-attention, which enable ViTs to learn better on small-size datasets~\cite{lee2021vision}.
\textbf{Details of all used ViTs are shown in Appendix}. For vanilla and EWA training, we always apply consistent CutMix, Mixup, Auto Augment, random erasing, and label smoothing for all models. On CIFAR-100, the input size is 32$\times$32, the patch size of ViT is set to 4, while 2 for PiT and Swin. On Tiny-ImageNet, the input size is 64$\times$64, the patch size is set to 8 for ViT, while 4 for PiT and Swin. All models are trained with the AdamW optimizer, the cosine schedule and a batch size of 128. For ViT-S, we train it for 300 epochs with 30 warm-up epochs, the learning rate is set to 0.0006, the weight decay is set to $0.06$. For the remaining models, we train them for 100 epochs with 10 warm-up epochs, the weight decay is set to 0.05, the learning rate is set to 0.003 (0.001 for Swin-T).

{\bfseries Results.} As shown in Tab.~\ref{table:2d training},  compared with vanilla training, EWA training can bring consistent performance improvement for various ViT architectures on different datasets. For standard ViT-S, EWA training can bring great improvements \{1.72\%, 2.09\%\} on \{CIFAR-100, Tiny-ImageNet\} respectively. 
Even for the specifically designed ViT architectures and their SL-enhanced counterparts, the addition of EWA training scheme can still improve their performance, which further confirms its effectiveness and generalization. For instance, SL-enhanced Swin-T with vanilla training has already achieved a relatively high top-1 accuracy, 80.31\% on CIFAR-100 and 64.34\% on Tiny-ImageNet. It can still obtain \{0.21\%, 0.79\%\} performance boosts from EWA training for each dataset.

\subsubsection{3D Vision tasks} \vspace{-1mm}
{\bf Settings.} To demonstrate the general training ability of our method, we also conduct experiments on point cloud classification (sparse) and semantic segmentation (dense) tasks. Following the experimental setting of PointBERT \cite{yu2022point}, we conduct the point cloud classification on the ModelNet40 \cite{qiu2021geometric}. Following the experimental setting of Pix4point \cite{qian2022pix4point}, we conduct the point cloud semantic segmentation task on the S3DIS \cite{xu2020grid} and leave area 5 as testing, other areas as training. Following PointBERT and Pix4point, we use the standard encoder in ViT-S as transformer backbone.

\begin{table*}[t]
\caption{Performance comparison between vanilla and EWA training on point cloud classification and point cloud semantic segmentation.}
\vspace{-1mm}
\label{table:3d}
    \centering
	\begin{minipage}{0.48\linewidth}
		\centering
        \subcaption{Classification on ModelNet40.}
\begin{tabular}{cccc}
\hline
Scheme & \begin{tabular}[c]{@{}c@{}} Acc.(w/o vote)\end{tabular} & \begin{tabular}[c]{@{}c@{}} Acc.(w vote)\end{tabular} \\ \hline
Vanilla & 92.42 & 92.67  \\
EWA    & \textbf{92.54}  & \textbf{92.79}    \\ \hline
\end{tabular}
	\end{minipage}
	\hfill
	\begin{minipage}{0.48\linewidth}
		\centering
        \subcaption{Semantic segmentation on S3DIS Area5.}
\begin{tabular}{cccc}
\hline
Scheme & \begin{tabular}[c]{@{}c@{}} OA \end{tabular} & \begin{tabular}[c]{@{}c@{}} mAcc. \end{tabular} & \begin{tabular}[c]{@{}c@{}}mIoU\end{tabular} \\ \hline
Vanilla & 89.53  & 72.43 &  66.62  \\
EWA & \textbf{90.05}  & \textbf{73.83} &  \textbf{68.36}  \\ \hline
\end{tabular}
\end{minipage}
\vspace{-3mm}
\end{table*}

{\bf Results.}
As shown in Tab.~\ref{table:3d}, the proposed EWA training scheme obtains consistent performance gain on both point cloud classification and semantic segmentation tasks. Notably, our method achieves an improvement of 0.52\% in overall accuracy (OA), 1.40\% in mean accuracy (mAcc) and 1.74\% in mean IoU (mIoU) on the segmentation task.

\subsection{EWA Fine-tuning} \vspace{-1mm}
{\bfseries Settings.} To evaluate EWA fine-tuning, we take pre-trained ViT-B and ViT-S from timm library and fine-tune them on CIFAR-100 and Food-101. For both CIFAR-100 and Food-101 \cite{kaur2017combining}, the batch size is set to 128 and all images are scaled to 224×224 for fine-tuning. For CIFAR-100, the total fine-tuning steps is set to 5000 with 500 warm-up steps. For Food-101, we first warm up 100 steps and the number of total steps is 1000. We use SGD optimizer with 0.9 momentum. The learning rate is 0.01 for CIFAR-100 and 0.03 for Food-101. We just simply place the MoE on every other layer and set the number of MoE layer's experts to 4. The $\beta$ of Experts Weights Averaging (Eq.~\ref{formulation:ewa}) is linearly increased with current step from 0 to the given hyperparameter $share\ rate$.

\begin{table*}[t]
\caption{Performance comparison of vanilla and EWA fine-tuning on various models and datasets.}
\vspace{-1mm}
\label{table:2d finetuning}
    \centering
	\begin{minipage}{0.48\linewidth}
		\centering
        \subcaption{on Food-101 dataset}
\begin{tabular}{cccc}
\hline
Model & \begin{tabular}[c]{@{}c@{}}Vanilla\\ Top-1\end{tabular} & \begin{tabular}[c]{@{}c@{}}EWA\\ Top-1\end{tabular} & \begin{tabular}[c]{@{}c@{}}$\Delta$\\ Top-1\end{tabular} \\ \hline
ViT-S & 88.04                                                   & 88.42                                               & +0.38                                             \\
ViT-B & 86.78                                                   & 87.32                                               & +0.54                                             \\ \hline
\end{tabular}
	\end{minipage}
	\hfill
	\begin{minipage}{0.48\linewidth}
		\centering
        \subcaption{on CIFAR-100 dataset}
\begin{tabular}{cccc}
\hline
Model & \begin{tabular}[c]{@{}c@{}}Vanilla\\ Top-1\end{tabular} & \begin{tabular}[c]{@{}c@{}}EWA\\ Top-1\end{tabular} & \begin{tabular}[c]{@{}c@{}}$\Delta$\\ Top-1\end{tabular} \\ \hline
ViT-S & 90.22                                                   & 90.67                                               & +0.45                                             \\
ViT-B & 90.71                                                   & 91.42                                               & +0.71                                             \\ \hline
\end{tabular}
\end{minipage}
\vspace{-2mm}
\end{table*}

{\bfseries Results.} Tab.~\ref{table:2d finetuning} shows the fine-tuned performance comparison between vanilla and EWA fine-tuning. Although the performance of vanilla fine-tuning is pretty high, EWA fine-tuning can still bring further improvements. Specifically, equipped with EWA fine-tuning, ViT-S and ViT-B obtain \{0.38\%, 0.45\%\} and \{0.54\%, 0.71\%\} performance boosts on \{Food-101, CIFAR-100\} respectively.

\subsection{Ablation Studies} \vspace{-1mm}
\begin{minipage}{0.49\linewidth}
{\bfseries Share schedule and Share rate.} We first study how to set the hyperparameter $share\ rate$ and share schedule for Experts Weights Averaging. In the above, the share schedule is set to linear increasing by default, we here compare it with the constant share schedule. 
For $share\ rate$, we adjust it  within \{0.1, 0.2, 0.3, 0.4, 0.5\}. 
We train ViT-S on CIFAR100 using EWA training with various share schedules and $share\ rate$. As shown in Fig.~\ref{fig:ablation}, EWA training with different schedules and $share\ rate$ can always outperform vanilla training, and EWA training with the linear increasing schedule performs better than the constant schedule under most cases. \textbf{Ablation studies about the number of experts are shown in the Appendix}.
\end{minipage}
\hfill
\begin{minipage}{0.49\linewidth}
\vspace{-2mm}
\begin{figure}[H]
\centering
\includegraphics[width=0.98\textwidth]{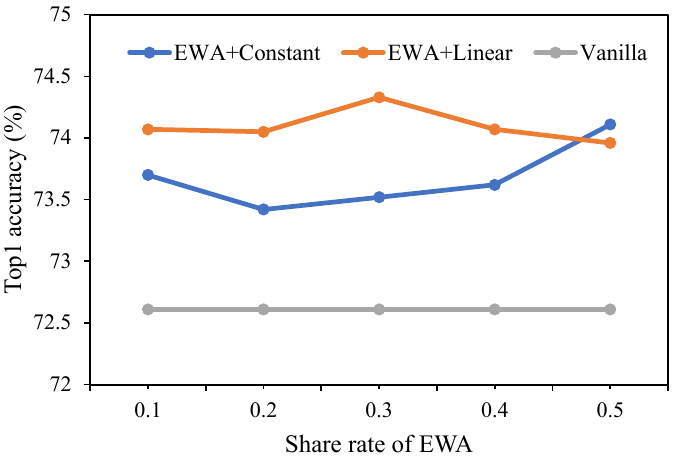}
\vspace{-1mm}
\caption{Ablation study on share schedules and share rates of the proposed EWA scheme.}
\label{fig:ablation}
\end{figure}
\end{minipage}

\section{More Discussions And Findings}
\begin{minipage}{0.49\linewidth}
More interestingly, we find that, on small 2D visual datasets and 3D visual tasks, naive MoE results in lower performance than its original dense counterparts, and our EWA technology can be seamlessly applied to naive MoE and help it learn better. \\ 
{\bfseries naive MoE.}
Following the settings of training ViT-S on CIFAR-100, Tiny-ImageNet, ModelNet40 and S3DIS, we further train our V-MoE-S, which replaces some FFN of ViT-S with naive top-1 routing MoE \cite{fedus2022switch} every other block. We set the number of experts as 4, the weight of load balance loss $\lambda$ as 0.01 and the capacity ratio $C$ as 1.05 for 2D datasets and 1.2 for 3D datasets. As shown in Tab.~\ref{table:early-ewa}, with vanilla training, naive V-MoE-S degrades performance compared to ViT-S on different datasets. \\
\end{minipage}
\hfill
\begin{minipage}{0.49\linewidth}
\vspace{-11mm}
\begin{table}[H]
\caption{Comparison between different models with different training schemes. Compared to ViT-S with vanilla training, V-MoE-S with vanilla training degrades the performance, while V-MoE-S with Early EWA improves the performance by a large margin, across 2D/3D different datasets.}
\vspace{1mm}
\label{table:early-ewa}
\centering
\setlength{\tabcolsep}{1.5mm}{
\begin{tabular}{cccc}
\hline
\begin{tabular}[c]{@{}c@{}}Model\\ +scheme\end{tabular} & \begin{tabular}[c]{@{}c@{}}ViT-S\\ +Vanilla\end{tabular} & \begin{tabular}[c]{@{}c@{}}V-MoE-S\\ +Vanilla\end{tabular} & \begin{tabular}[c]{@{}c@{}}V-MoE-S\\ +Early EWA\end{tabular} \\ \hline
\begin{tabular}[c]{@{}c@{}}C100\\ Acc\end{tabular}      & 72.61                                                    & 72.17                                                      & \textbf{74.31}                                               \\ \hline
\begin{tabular}[c]{@{}c@{}}Tiny-Img\\ Acc\end{tabular}  & 57.41                                                    & 56.28                                                      & \textbf{59.91}                                               \\ \hline
\begin{tabular}[c]{@{}c@{}}MN40\\ Acc\end{tabular}      & 92.42                                                    & 92.27                                                      & \textbf{92.63}                                               \\ \hline
\begin{tabular}[c]{@{}c@{}}S3DIS\\ mIoU\end{tabular}    & 66.62                                                    & 66.36                                                      & \textbf{67.67}                                               \\ \hline
\end{tabular} }
\end{table}
\end{minipage}

{\bfseries naive MoE + Early EWA.} Things change when we introduce Experts Weights Averaging (EWA) into naive MoE. Considering that using EWA throughout the entire training process will finally lead to nearly identical weights across experts, we only perform EWA on naive MoE in the early stage of training. By default, we use constant share schedule and only perform Early EWA in the first half of total training epochs. With Early EWA training, the performance of V-MoE-S is greatly improved, resulting in a top-1 accuracy gain of \{2.14\%, 3.63\%\} on \{CIFAR-100, Tiny-ImageNet\} respectively. In addition, our Early EWA training can improve the performance of V-MoE-S on various 3D vision tasks, with 0.36\% accuracy gain on ModelNet40 classification, and 0.30\% OA gain, 1.40\% mAcc gain, 1.31\% mIoU gain on S3DIS Area5 segmentation. \textbf{Further studies are shown in Appendix.}

\section{Conclusion}
In conclusion, our proposed training scheme for ViTs achieves performance improvement without increasing inference latency and parameters. By designing efficient MoEs and EWA during training, and converting each MoE back into a FFN by averaging the experts during inference, we decouple the training and inference phases of ViTs. Comprehensive experiments across various 2D and 3D visual tasks, ViT architectures, and datasets demonstrate the effectiveness and generalizability. The theoretical analysis further supports our approach, and our training scheme can also be applied to fine-tuning ViTs and improve the effectiveness of naive MoE in various visual tasks. Overall, our proposed training scheme provides a promising direction for enhancing the performance of ViTs in various visual tasks, and we believe it could lead to further research and development in this field.

{
\small
\bibliographystyle{plain}
\bibliography{neurips_2023}
}

\end{document}